\documentclass{article}

\usepackage{arxiv}
\PassOptionsToPackage{numbers}{natbib}
\usepackage[utf8]{inputenc} 
\usepackage[T1]{fontenc}    
\usepackage{hyperref}       
\usepackage{url}            
\usepackage{booktabs}       
\usepackage{amsmath,amssymb,amsfonts}
\usepackage{nicefrac}       
\usepackage{microtype}      
\usepackage{lipsum}		
\usepackage{graphicx}
\usepackage{natbib}
\usepackage{doi}
\usepackage{tabularx}
\usepackage{textcomp}
\usepackage{hyperref} 
\usepackage{algorithmic}

\title{Res-VMamba:  Fine-Grained Food Category Visual Classification Using Selective State Space Models with Deep Residual Learning}

\author{ \href{https://orcid.org/0000-0003-0807-0217}{\includegraphics[scale=0.06]{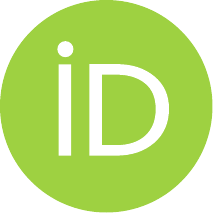}\hspace{1mm}Chi-Sheng Chen\thanks{Corresponding author} \thanks{These authors’ contributions are equal.}} \\
Graduate Institute of \\ Biomedical Electronics and Bioinformatics \\
 National Taiwan University\\
	No.1, Sec.4, Roosevelt Road, Taipei, 10617, Taiwan \\
	\texttt{m50816m50816@gmail.com} \\
	\And
	\href{https://orcid.org/0009-0007-2725-1239}
 {\includegraphics[scale=0.06]{orcid.pdf}\hspace{1mm}Guan-Ying Chen\thanks{These authors’ contributions are equal.}} \\
  Graduate Institute of \\ Biomedical Electronics and Bioinformatics \\
 National Taiwan University\\
	No.1, Sec.4, Roosevelt Road, Taipei, 10617, Taiwan \\
	\texttt{d03945004@g.ntu.edu.tw} \\
 \And
	\href{https://orcid.org/0009-0000-5381-2625}
 {\includegraphics[scale=0.06]{orcid.pdf}\hspace{1mm}Dong Zhou} \\
	University of Liverpool\\
	L69, 3BX, United Kingdom \\
	\texttt{Dong.zhou93@gmail.com} \\
 \And
	\href{https://orcid.org/0000-0002-3345-168X}
 {\includegraphics[scale=0.06]{orcid.pdf}\hspace{1mm}Di Jiang} \\
       Department of Otolaryngology \\
	Dongguan People’s Hospital\\
	No. 3 South, Xinganyong Wan Road, \\Dongguan City, Guangdong Province, China \\
	\texttt{shaouzhixing@163.com} \\
  \And
	\href{https://orcid.org/0009-0006-2740-8251}
 {\includegraphics[scale=0.06]{orcid.pdf}\hspace{1mm}Dai-Shi Chen\thanks{Corresponding author}} \\
       Department of Otolaryngology \\
	Shenzhen People’s Hospital\\
	No. 1017, Dongmen North Road, \\  Shenzhen City, Guangdong Province, China \\
	\texttt{cdsginqtbing@hotmail.com} \\
}

\renewcommand{\shorttitle}{\textit{arXiv} Template}

\hypersetup{
pdftitle={A template for the arxiv style},
pdfsubject={cs.CV},
pdfauthor={Chi-Sheng ~Chen, Guan-Ying ~Chen, Dong ~Zhou, Dai-Shi ~Chen},
pdfkeywords={Deep learning, fine-grained visual classification, food category recognition, state space model, mamba},
}

\begin{document}
\maketitle

\begin{abstract}
	Food classification is the foundation for developing food vision tasks and plays a key role in the burgeoning field of computational nutrition. Due to the complexity of food requiring fine-grained classification, recent academic research mainly modifies Convolutional Neural Networks (CNNs) and/or Vision Transformers (ViTs) to perform food category classification. However, to learn fine-grained features, the CNN backbone needs additional structural design, whereas ViT, containing the self-attention module, has increased computational complexity. In recent months, a new Sequence State Space (S4) model, through a Selection mechanism and computation with a Scan (S6), colloquially termed Mamba, has demonstrated superior performance and computation efficiency compared to the Transformer architecture. The VMamba model, which incorporates the Mamba mechanism into image tasks (such as classification), currently establishes the state-of-the-art (SOTA) on the ImageNet dataset. In this research, we introduce an academically underestimated food dataset CNFOOD-241, and pioneer the integration of a residual learning framework within the VMamba model to concurrently harness both global and local state features inherent in the original VMamba architectural design. The research results show that VMamba surpasses current SOTA models in fine-grained and food classification. The proposed Res-VMamba further improves the classification accuracy to 79.54\% without pretrained weight. Our findings elucidate that our proposed methodology establishes a new benchmark for SOTA performance in food recognition on the CNFOOD-241 dataset. The code can be obtained on GitHub: \underline{https://github.com/ChiShengChen/ResVMamba}.
\end{abstract}


\section{Introduction}
Food plays a significant role in human life, and with the advent of modern technology and dramatically changed lifestyle and dietary habits, there has been a growing focus on food computing \cite{14}. Within the realm of food computing, food recognition is an important research direction in computer vision and machine learning. In the industry, food image classification can be applied to automating restaurant cooking processes, self-checkout systems, and kitchen waste management. Additionally, food recognition is a vital component in numerous health-related applications, encompassing the nutritional analysis of food and dietary habit management. However, despite ongoing research, there are still several challenges in food image classification. 

In addition to the impacts of the photo-captured environment, noise in images, and image quality, the biggest challenge for food classification is low inter-class variation and high intra-class variation. The same food category can appear differently depending on cooking methods, seasonings, plating styles, and other preparation factors. While different food categories, even subtle differences in ingredients can result in visually similar but semantically different food types, such as shredded pork fried rice and shrimp fried rice. Addressing these issues requires techniques capturing fine-grained features to distinguish between food classes.

Fine-grained visual classification represents a formidable task within the field of computer vision, seeking to identify various subcategories within a broader category, such as the various species of birds \cite{1}, \cite{13}, medical images \cite{2}, cars \cite{3}, \cite{4}, aircraft \cite{5}, dogs \cite{6}, pets \cite{7}, flowers \cite{8}, natural images \cite{9}, \cite{10}, etc. Food image recognition is also an important branch of Fine-Grained Visual Classification (FGVC) \cite{11}, \cite{12}. The pivotal point in the fine-grained classification is that in addition to learning global features, the model should be able to integrate local features to get global-local information mutually in order to achieve better recognition capabilities. Current methods can be summarized as developing model subnet for localization features or delivering better feature learning methods.

Recently, models based on the State Space Model, such as VMamba \cite{43}, are regarded as outperforming ViTs on large image datasets like ImageNet \cite{42}. It retains the advantage of capturing both local and global information from input images as ViTs while also enhancing the model speed. However, there is still a lack of research on the application of VMamba to fine-grained datasets. Therefore, this study endeavors to employ VMamba on food images and introduces a model, Res-VMamba, designed specifically for fine-grained datasets. This model incorporates a mechanism to share global and local feature states, aiming to enhance performance on detailed image classification tasks.

A well-defined dataset significantly influences the development of possible research topics and the feature-learning capabilities of models. In this study, we utilized the CNFOOD241 dataset \cite{12}, \cite{65}. CNFOOD241 is a Chinese food dataset created by expanding ChineseFoodNet \cite{32}, including correcting incorrect labels and increasing the number of images and food categories. In addition to model training, we provided a comparative analysis of CNFOOD241 and other food datasets, illustrating its suitability for research. Unlike other food databases, CNFOOD241 preserves the aspect ratio of images and standardizes the size to 600$\times$600 pixels. This preprocessing step prevents image deformation during data augmentation, which could potentially lead to models learning incorrect semantic features. Furthermore, CNFOOD241 exhibits the relative imbalanced data distribution, making it a more challenging dataset for experimentation. Considering the above factors, we selected CNFOOD241 for experiments on fine-grained food classification.

The contributions of this work are stated as follows:
\begin{itemize}
    \item We provide comparative studies on the food dataset and clarify the research value of CNFOOD241. To enhance the rigor of our study, we have further partitioned the dataset into separate test and validation segments as a new fine-grained image classification benchmark.
    \item We first introduce the state space model into fine-grained image classification, and the proposed Res-VMamba outperforms state-of-the-art approaches on the CNFOOD-241 dataset.
\end{itemize}

\begin{table}
    \centering
    \begin{tabular}{ccccc}
       \hline
        Dataset& Year&Classes/Images&Type&Public\\
        \hline
        PFID \cite{16} &2009 &101/4,545&Western&$\times$\\
        Food50 \cite{17} &2010&50/5,000&Misc.&$\times$\\
        Food85 \cite{18}&2010&85/8,500&Misc.&$\times$\\
        UEC Food100 \cite{19}&2012& 100/14,361&Japanese&$\surd$\\
        UEC Food256 \cite{20}&2014& 256/25,088&Japanese&$\surd$\\
        ETH Food-101 \cite{21}&2014& 101/101,000&Western&$\surd$\\
        Diabetes \cite{22} &2014&11/4,868&Misc.&$\times$\\
        UPMC Food-101 \cite{23}&2015&101/ 90,840&Western&$\surd$\\
        Geo-Dish \cite{24}&2015&701/117,504&Misc.&$\surd$\\
        UNICT-FD889 \cite{25}&2015&889/3,583&Misc.&$\surd$\\
        Vireo Food-172 \cite{26}&2016&172/110,241&Chinese&$\surd$\\
        Food-975 \cite{27}&2016 & 975/37,785&Misc.&$\times$\\
        Food500 \cite{28} &2016& 508/148,408&Misc.&$\times$\\
        Food11 \cite{11} &2016&11/16,643&Misc.&$\surd$\\
        UNICT-FD1200 \cite{30} &2016& 1,200/4,754&Misc.&$\surd$\\
        Food524DB \cite{31}&2017&524/247,636&Misc.&$\times$ \\
        ChineseFoodNet \cite{32}&2017&208/192,000&Chinese&$\surd$ \\
        Vegfru \cite{33}&  2017&292/160,000&Misc.&$\surd$\\
        Sushi-50 \cite{34}&  2019&50/3,963&Japanese&$\surd$\\
        FoodX-251 \cite{35}&  2019&251/158,846&Misc.&$\surd$\\
        ISIA Food-200 \cite{36}&  2019&200/197,323&Misc.&$\surd$\\
        FoodAI-756 \cite{37}&  2019&756/400,000&Misc.&$\times$\\
        Taiwanese-Food-101 \cite{38}&2020&101/20200&Chinese&$\surd$\\
        ISIA Food-500 \cite{39}&  2020&500/399,726&Misc.&$\surd$\\		
        Food2K \cite{40}&2021&2,000/1,036,564&Misc.&$\surd$\\
        MyFoodRepo-273 \cite{41} &2022& 273/24,119&Misc.&$\times$\\
        \hline
        \textbf{CNFOOD-241 \cite{65}}&\textbf{2022}&\textbf{241/191,811}&\textbf{Chinese}&$\surd$\\
        \hline
    \end{tabular}
    \caption{Comparison of current food recognition datasets.}
    \label{tab:my_label}
\end{table}

\section{Related work}
\label{sec:headings}


\subsection{Food Recognition Datasets}
In the burgeoning field of food computation, the proliferation of food datasets has marked a significant advancement, drawing widespread academic and practical interest. From the inception of datasets like ETH Food-101 \cite{21}, which introduced over one hundred thousand images of Western food varieties, to the expansive collections of ISIA Food-500 \cite{39} and Food2K \cite{40}, encompassing nearly four hundred thousand and over a million images respectively, the evolution is notable. These datasets, predominantly sourced through web scraping, have been instrumental in advancing computational gastronomy and nutrition studies. However, they share a critical limitation: a lack of uniformity in the size distribution of images across different categories. This variance can lead to substantial discrepancies in some categories, where a few images might significantly exceed the average size of others, potentially skewing the dataset's overall utility and introducing biases in the processing and classification results obtained after resizing images for analysis.

The issue of image size inconsistency poses challenges in maintaining the accuracy and reliability of computational models, especially those reliant on CNNs, ViTs and other image-processing architectures designed to extract detailed features from visual inputs. As depicted in Table 2, the disparity in image sizes may affect the performance of these models, leading to deviations in the extracted category-specific information and potentially impacting the overall effectiveness of the computational analysis.

In response to these challenges, our search for a more consistent and high-resolution dataset led us to CNFOOD-241. Among publicly available food datasets with uniform image sizes, such as UNICT-FD889 \cite{25}, Vireo Food-172 \cite{26}, and UNICT-FD1200 \cite{30}, CNFOOD-241 distinguishes itself by offering the highest resolution. This characteristic renders it an exceptional resource for conducting detailed image analyses within the food computation domain, facilitating more accurate and reliable studies in food recognition, nutritional analysis, and other related areas. The unveiling of the CNFOOD-241 dataset marks a significant advancement in fulfilling the essential demand for high-quality, uniform datasets within the domain of food computation, thereby facilitating novel pathways for research and innovation. Experimental evidence further indicates that this dataset presents a considerable challenge, enhancing its benchmark value and making it a critical asset for advancing the state of the art in food recognition technologies. This complexity not only tests the limits of current models but also underscores the dataset's potential as a valuable tool for benchmarking advancements in the field.

\subsection{Fine-Grained Visual Classification}
In the domain of computer vision, food recognition is categorically placed within the realm of FGVC, a field distinguished by its focus on distinguishing between closely related subcategories within a broader category. This area has seen the development and application of several state-of-the-art (SOTA) models, each contributing to advancements in dataset-specific performance. Among these, the CMAL-Net \cite{44} stands out, having been constructed by integrating three expert modules with a CNN-based backbone. Each expert module processes feature maps from specific layers, delivering both a categorical prediction and an attention region. This attention region not only highlights areas of interest within the images but also serves as a means for data augmentation for the other expert modules, thereby enhancing the model's overall accuracy and robustness.

Furthermore, EfficientNet \cite{45}, a model that has retained its relevance over time, employs Network Architecture Search (NAS) to devise a set of rules that are consolidated into Compound Model Scaling. This approach strategically scales model architecture to optimize performance. Additionally, ConvNeXT \cite{46}, inspired by the Swin Transformer \cite{53} architecture, reimagines CNNs to surpass the Swin Transformer's performance on ImageNet, marking a significant achievement in model design.

Another noteworthy development is the introduction of the Squeeze-and-Excitation (SE) block into ViT, leading to the creation of RepViT \cite{47}. This model emerges as a successor in the ViT era following the SENet \cite{48}, demonstrating impressive performance in image recognition tasks. Traditional ViTs, which marked a pivotal moment by outperforming models with purely CNN-based backbones in image recognition domains, further underscore the ongoing evolution and innovation within FGVC. These advancements not only push the boundaries of what's achievable in fine-grained classification but also provide a rich tapestry of methodologies for addressing the nuanced challenges of food recognition.

\subsection{Food Image Recognition}
Early food recognition systems primarily used traditional machine learning algorithms. Researchers extract handcrafted features from images using methods such as color histograms \cite{67}, Scale-Invariant Feature Transform (SIFT), Histogram of Oriented Gradients (HOG) \cite{66}, Gabor textures \cite{67}, and Local Binary Pattern (LBP) \cite{66}. These extracted features were then fed into classifiers such as SVM \cite{72} for categorization. While achieving reasonable performance, these early methods relied heavily on manual feature engineering and were limited by image quality and variability in food appearances. 

The emergence of deep learning revolutionized food identification research. Researchers began applying convolutional neural networks, such as AlexNet \cite{68}, ResNet-50 \cite{69}, EfficientNet \cite{70}, and Inception v3 \cite{71}, to food image data using transfer learning methods. This approach eliminated the need for manual feature extraction and allowed models to learn hierarchical visual representations from large-scale labeled datasets. Subsequent research enhanced food recognition capabilities using ensemble networks, multi-task learning, and other techniques. Notably, PRENet \cite{40} emerged as a milestone in food recognition, integrating three different branches, each tailored for capturing different aspects of food images. By fusing features from low and high-level layers, PRENet achieved SOTA performance on CNN models.

Recently, ViT has have gained popularity in food image analysis due to their ability to capture long-range dependencies. ViT models divide images into patches and represent them as sequence data, applying self-attention to capture relationships between patches. Ongoing research continues to integrate ViT with data augmentation, semi-supervised learning, multi-model fusion, and other techniques, pushing the boundaries of food understanding from images. Our research pioneers the application of the State Space Model to food recognition and aims to bring breakthroughs in this field. 

\subsection{State Space Model on Visual Recognition}

Recent research predominantly utilizes CNNs and ViTs for the task of classifying food categories. However, the capability of these models to detect features has not reached the SOTA in large-scale image classification challenges lately, being outperformed by a new generation of models known as Structured State Space for Sequence (S4) \cite{54} modeling. The improvement of S4 models with a Selection mechanism and their execution using a Scan (S6) \cite{55}, informally known as Mamba, has shown to outclass the Transformer architecture in handling long sequences. The VMamba \cite{43} model, which applies the Mamba mechanism to downstream image tasks like classification, now sets the SOTA benchmark on the ImageNet dataset. There are several Mamba models used on vision related task such as VMamba, Vision Mamba \cite{56} have tried to use Mamba to do visual downstream tasks like image classification and object detection, but Vision Mamba more focus on inference speed and GPU-memory usage efficiency. U-Mamba \cite{57}, VM-UNet \cite{58}, Swin-UMamba \cite{59}, nnMamba \cite{60} and MambaMorph \cite{61} have replaced convolutional blocks or downsample blocks to Mamba blocks applied on medical image segmentation tasks. Vivim \cite{62} used Mamba-based backbone for medical video object segmentation. However, there is still lack of utiling VMamba on eithor fine-grained data or food recognition task, in this work we propose the VMamba-based model into these downstream task.
\begin{figure*}[t]
\centering
\includegraphics[scale=0.4]{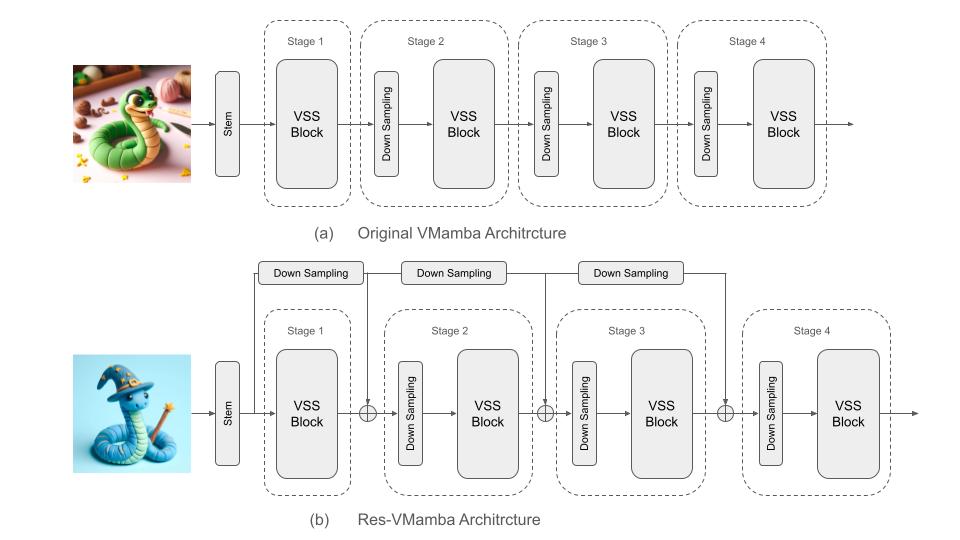}
\caption{The comparison between VMamba and our proposed Res-VMamba.}
\label{fig:cnfood-241-report-new}
\end{figure*}
\subsection{Deep Residual Learning on Space State Model}
Deep residual learning is a technique employed in training deep neural networks that offers several notable advantages introduced by ResNet \cite{51}. Its primary advantage lies in addressing challenges such as vanishing gradients and exploding gradients, which commonly impede the training of deep networks. By introducing residual blocks and skip connections, deep residual learning facilitates the flow of gradients throughout the network, effectively mitigating the issue of vanishing gradients. Consequently, it enables the training of deeper neural networks without sacrificing performance. Deeper networks afford the extraction of more complex features, thereby enhancing the model's representational capacity. Additionally, the training process in deep residual learning converges more efficiently due to expedited gradient propagation via skip connections, resulting in reduced training time and computational costs. Moreover, the technique enhances the model's generalization capabilities by effectively fitting training data while mitigating the risk of overfitting. 

However, to our best knowledge, there is lack of research that use residual learning on VMamba. Hence, we want to introduce the residual learning structure into VMamba model.

\section{Methodology}
\label{sec:others}

In this section, we first introduce the preliminary knowledge of VMamba, then proposed the detail to our Res-VMamba structure. 
\subsection{State Space Models}
State Space Models (SSMs) are widely recognized as linear systems with time-invariance properties, mapping an input \( x(t) \in \mathbb{R}^L \) to an output \( y(t) \in \mathbb{R}^L \). These systems are mathematically formulated as linear ordinary differential equations (ODEs), as depicted in Equation (1), where the model's parameters are denoted by \( A \in \mathbb{C}^{N \times N} \), \( B \in \mathbb{C}^{N \times L} \), for a system state of dimension \( N \), and the direct link, \( D \in \mathbb{C}^L \). The state's derivative and output signals are described by the following equations:
\begin{equation}
\begin{aligned}
& h'(t) = Ah(t) + Bx(t) \\
& y(t) = Ch(t) + Dx(t)
\end{aligned}
\end{equation}

\subsection{Discretization}
When integrated into deep learning algorithms, State Space Models (SSMs), inherently continuous-time constructs, present substantial challenges. The discretization process is thus imperative. 

The primary aim of discretization is to transmute the continuous ODE into a discrete function. This conversion is vital for aligning the model with the input data's sample rate, thereby enabling computationally efficient operations \cite{63}. Given the input \( x_k \in \mathbb{R}^{L \times D} \), which is a sampled vector from the signal sequence of length \( L \), the ODE \cite{64} (Eq.1) can be discretized employing the zeroth-order hold approach:

\begin{equation}
\begin{aligned}
h_k &= A_d h_{k-1} + B_d x_k, \\
y_k &= C_d h_k + D x_k,
\end{aligned}
\label{discreteSystem}
\end{equation}

where \( A_d = e^{A\Delta} \), \( B_d = (e^{A\Delta} - I)A^{-1}B \), and \( C_d = C \), with \( B, C \in \mathbb{R}^{D \times N} \) and \( \Delta \in \mathbb{R}^D \). Following the practice in \cite{55}, the approximation of \( B \) through first-order Taylor series is refined as:

\begin{equation}
\bar{B} = (e^{A\Delta} - I)A^{-1}B \approx (\Delta A)(\Delta A)^{-1} \Delta B = \Delta B
\label{taylorApprox}
\end{equation}

\subsection{2D Selective Scan Mechanism}
The VMamba model introduces a novel Selective Scan Mechanism (S6), diverging from traditional Linear Time-Invariant (LTI) State Space Models (SSMs). This S6 mechanism, central to the VMamba framework, incorporates matrices \( B \in \mathbb{R}^{B \times L \times N} \), \( C \in \mathbb{R}^{B \times L \times D} \), and \( \Delta \in \mathbb{R}^{B \times L \times D} \), extracted from the input data \( x \in \mathbb{R}^{B \times L \times D} \), to imbue the system with contextual responsiveness and weight dynamism.

Furthermore, the Cross-Scan Module (CSM) is introduced to enhance spatial integration across the image. It unfolds image patches into sequences along rows and columns, and performs scanning across four directions, thereby enabling any pixel to integrate information from all others in different trajectories. These sequences are then reconfigured into a single image, culminating in a merged, information-rich new image.

\label{sec:guidelines}
\begin{figure}
    \centering
    \includegraphics[width=0.3\linewidth]{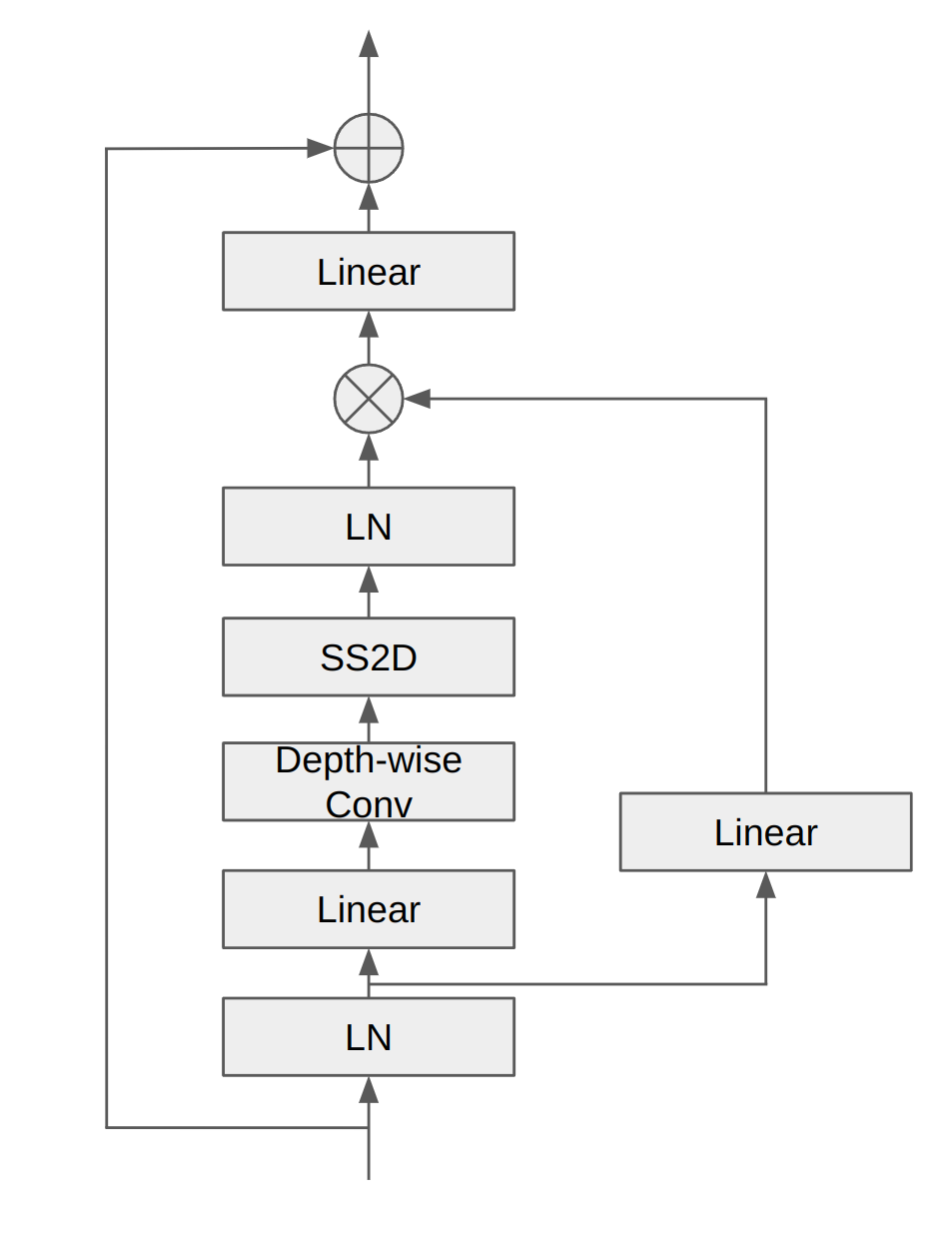}
    \caption{VSS Block.}
    \label{fig:vss_block}
\end{figure}
\subsection{VMamba Model}
The overall architecture of VMamba Model is illustrated in \cite{43}. The VMamba architecture, showcased in Figure~\ref{fig:cnfood-241-report-new}, commences by partitioning the input image into patches through a stem module, emulating ViTs while maintaining the 2D structure and translating the patches into a 1D sequence. This approach yields a feature map with dimensions \( \frac{H}{4} \times \frac{W}{4} \times C_1 \).

VMamba layers a series of VSS blocks like Figure~\ref{fig:vss_block} atop this feature map to construct ``Stage 1,'' preserving its dimensions. Hierarchical structures in VMamba are established via down-sampling in ``Stage 1'' through a patch merging process, as described in \cite{43}. More VSS blocks are then integrated, reducing the output resolution to \( \frac{H}{8} \times \frac{W}{8} \) for ``Stage 2.'' This down-sampling is reiterated to form ``Stage 3'' and ``Stage 4,'' with resolutions of \( \frac{H}{16} \times \frac{W}{16} \) and \( \frac{H}{32} \times \frac{W}{32} \), respectively. The resulting hierarchical design mirrors the multi-scale representation characteristic of renowned CNN models and some ViTs. Thus, VMamba's architecture emerges as a comprehensive and versatile candidate for a variety of vision-related applications with analogous requirements.

\subsection{Res-VMamba Model}
Inpired by ResNet, we proposed a new type of VMamba model, Res-Vmamba, a Mamba with residual learning mechanism. Res-VMamba architecture, as illustrated, is an advanced model configuration designed for efficient processing within the realm of computer vision. This architecture begins with a stem module that processes the input image, which is then followed by a series of VSS Blocks, arranged sequentially across four distinct stages.

Distinct from the original VMamba framework, the Res-VMamba architecture not only employs the VMamba structure as its backbone but also integrates raw data directly into the feature map. In order to distinguish it from the residual structure in the VSS block, we called that global-residual mechanism. This integration is anticipated to facilitate the sharing of global image features in conjunction with the information processed through the VSS blocks. The intention behind this design is to harness both the localized details captured by individual VSS blocks and the overarching global features inherent in the unprocessed input, thereby enriching the model's representational capacity and enhancing its performance on tasks requiring a comprehensive understanding of the visual data.

\section{Experiments}
\label{sec:guidelines}
\subsection{Dataset}
\begin{figure}
    \centering
    \includegraphics[width=0.5\linewidth]{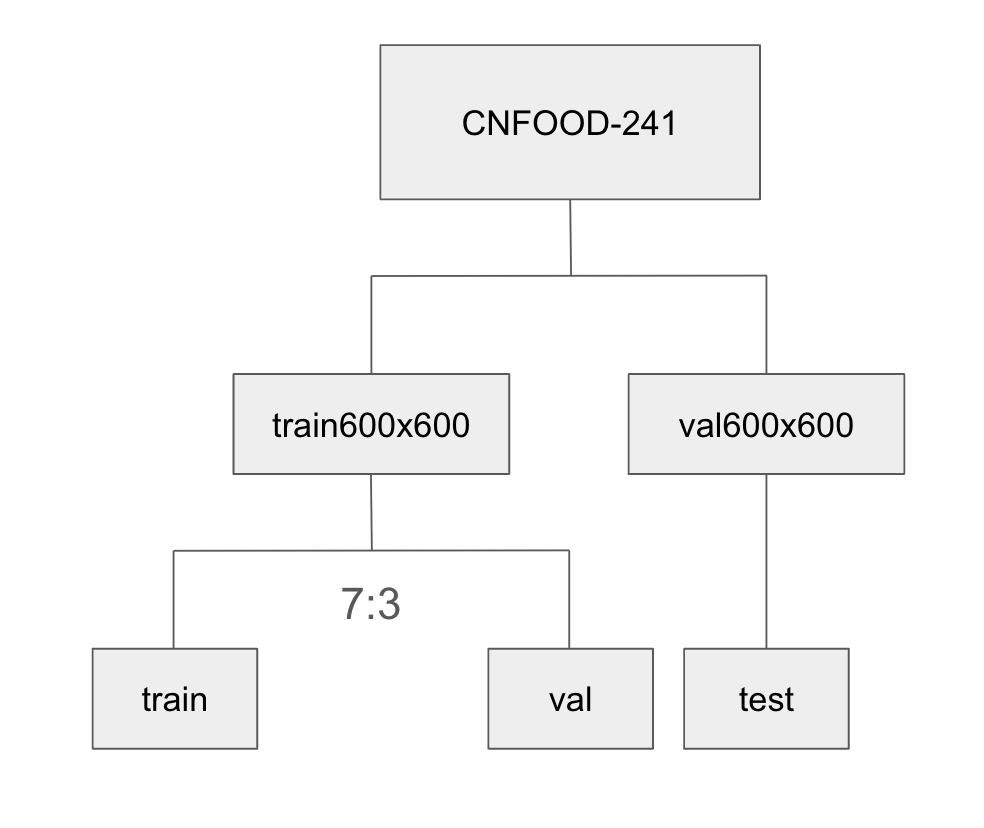}
    \caption{CNFOOD-241 data split flow.}
    \label{fig:dataset_split}
\end{figure}
In this work, to be fair to evaluate the models' performance, we used the CNFOOD-241 as the dataset for it has some awesome property as below: 

\subsubsection{Balanced Distribution on Image Sizes cross Categories}
{The CNFOOD-241 dataset possesses the largest (almost two hundred thousand images) uniform-sized (600$\times$600) image collection among publicly available food datasets.
 }

\subsubsection{Unbalanced Number of Images cross Categories on CNFOOD-241}
{
In order to evaluate the unbalance of image number in each category  through all the dataset, we calculate the normalized entropy. To normalize the entropy, we divide the entropy \(H\) by \(\log_2(n)\), where \(n\) is the total number of categories. This normalized entropy, denoted as \(H_{\text{norm}}\), is calculated as follows:
\begin{equation}
\begin{aligned}
H_{\text{norm}} = \frac{H}{\log_2(n)} 
\end{aligned}
\label{entropy}
\end{equation}

Given the original definition of entropy:
\begin{equation}
\begin{aligned}
 H = -\sum_{i=1}^{n} p_i \log_2(p_i) 
\end{aligned}
\label{entropy2}
\end{equation}

The normalized entropy \(H_{\text{norm}}\) ranges from 0 to 1, where
\(H_{\text{norm}} = 1\) indicates a perfectly balanced dataset, with each category having an equal share of the data.
\(H_{\text{norm}} = 0\) indicates complete imbalance, where all instances belong to a single category. This normalization allows for an easier comparison of entropy values across datasets with different numbers of categories, providing a standardized measure of category balance. The entropy result is shown as Table~\ref{tab:class_entropy}, we observe that CNFOOD-241 exhibits a greater degree of class imbalance compared to other datasets, which consequently enhances the challenging nature of this dataset.

Moreover, to establish CNFOOD-241 as a more equitable benchmark, we partitioned the dataset into training, validation, and test subsets at random. The training and validation sets were divided in a 7:3 ratio, respectively, through a randomized selection process. Finally, the training set contains 119,514 images and the validation set has 51,354 images in all 241 categories. The testing set is the original
 'val600x600' folder in CNFOOD-241 dataset, which include 20,943 images as Figure~\ref{fig:dataset_split}. The train-validation split list can be available on the github: \underline{https://github.com/ChiShengChen/ResVMamba}.
}
\begin{table*}
    \centering
    \begin{tabularx}{\linewidth}{@{}lXXXXl@{}}
       \hline
        Dataset& Max. H &Min. H & Max. W &Min. W & Mean $\pm$ Std. H$\times$W\\
        \hline
        UEC Food100 &800&71& 800 & 80&358.25$\pm$139.46$\times$457.40$\pm$180.58\\
        UEC Food256 &800&71& 800 & 80&406.99$\pm$117.16$\times$492.98$\pm$136.71\\
        ETH Food-101 &512&122& 512 & 193&475.37$\pm$65.31$\times$495.79$\pm$45.67\\
        UPMC Food-101&2960&120&2000 & 120&459.13$\pm$225.54$\times$559.80$\pm$268.67\\
        Geo-Dish&-*&-*&-*&-*&-*\\
        UNICT-FD889&240&240&320 & 320&240$\pm$0.00$\times$320$\pm$0.00\\
        Vireo Food-172&256&256&256&256&256$\pm$0.00$\times$256$\pm$0.00\\
        Food11 &6144&207& 9542&220&496.01$\pm$180.38$\times$531.55$\pm$250.92\\
        UNICT-FD1200 &240 &240&320 & 320&240$\pm$0.00$\times$320$\pm$0.00\\
        ChineseFoodNet&-*&-*&-*&-*&-* \\
        Vegfru&  6016&45&7360&48&309.73$\pm$272.46$\times$374.52$\pm$338.21\\
        Sushi-50&  4752&232&6570&285&767.20$\pm$584.50$\times$955.80$\pm$758.16\\
        FoodX-251&  2744&256&2733&256&287.48$\pm$75.45$\times$341.24$\pm$86.28\\
        ISIA Food-200&  8868&70&10030&78&707.28$\pm$561.74$\times$830.91$\pm$654.40\\
        Taiwanese-Food-101 &7360&61&6720&83&613.76$\pm$372.83$\times$771.91$\pm$455.90\\
        ISIA Food-500&  8868&100&10361&78&700.78$\pm$566.41$\times$822.08$\pm$667.43\\		
        Food2K&1489&75&1717&97&381.26$\pm$74.96$\times$ 515.89$\pm$102.72\\
        \hline
        \textbf{CNFOOD-241}&600&\textbf{600}&600&\textbf{600}&\textbf{600$\pm$0.00$\times$600$\pm$0.00}\\
        \hline
    \end{tabularx}
    \caption{Image size statistics comparison of current open datasets of food recognition (pixels). Max., Min., Std. are maximum, minimum and standard deviation respectively. H, W denotes the height and width of image respectively. The CNFOOD-241 dataset has the largest balenced image size in all the food image dataset we were available. *The data download link on the Geo-Dish (http://isia.ict.ac.cn/dataset/Geolocation-food/) and ChineseFoodNet (https://goo.gl/kWNV8a) official pages are dead, so we can not get the dataset to analyze image properties.}
    \label{tab:my_label}
\end{table*}

\begin{table}
    \centering
    \begin{tabular}{ccccc}
       \hline
        Dataset& Category Entropy\\
        \hline
        UEC Food100 &0.9758\\
        UEC Food256 &0.9891\\
        ETH Food-101 &\textbf{1.0000}\\
        UPMC Food-101&0.9999\\
        UNICT-FD889&0.9832\\
        Vireo Food-172&0.9876\\
        Food11 &0.9565\\
        UNICT-FD1200 &0.9883\\
        Vegfru&  0.9759\\
        Sushi-50& 0.9951\\
        FoodX-251&  0.9974\\
        ISIA Food-200&  0.9889\\
        Taiwanese-Food-101 &\textbf{1.0000}\\
        ISIA Food-500& 0.9880\\		
        Food2K&0.9821\\
        CNFOOD-241&0.9780\\
        \hline
    \end{tabular}
    \caption{Comparison of entropy on current food recognition datasets.}
    \label{tab:class_entropy}
\end{table}

\subsection{Performance Metrics}
The performance of the models are evaluated by the top-k accuracy. 
Top-k accuracy is defined as the proportion of test samples for which the correct label is among the top \(k\) labels predicted by the model. Mathematically, it can be expressed as:

\begin{equation}
\begin{aligned}
 \text{Top-k Accuracy} = \frac{1}{N} \sum_{i=1}^{N} \mathbb{1}\left(y^{(i)} \in P_k^{(i)}\right) 
 \end{aligned}
\label{topk}
\end{equation}

where:
\begin{itemize}
    \item \(N\) is the total number of samples in the test set.
    \item \(y^{(i)}\) is the true label for the \(i\)-th sample.
    \item \(P_k^{(i)}\) is the set of top \(k\) predictions made by the model for the \(i\)-th sample.
    \item \(\mathbb{1}\) is the indicator function, which is 1 if \(y^{(i)} \in P_k^{(i)}\) (the true label is among the top \(k\) predictions) and 0 otherwise.
\end{itemize}

This metric is particularly useful for evaluating models on tasks where the goal is to provide a set of potential labels for each input and the exact rank within the top \(k\) is not critically important.

\subsection{Implementation Details}\label{formats}
 Adhering to the protocol delineated in \cite{43}, Res-VMamba embarks on a comprehensive training regimen on CNFOOD-241, the backbone uses the VMamba-S, extending over 150 epochs with an initial warmup period covering the initial 20 epochs, and leverages a batch size of 128. The training schema integrates the AdamW optimizer, with the beta parameters set at \( (0.9, 0.999) \), and momentum fixed at 0.9. A cosine decay schedule modulates the learning rate, commencing with an initial learning rate of \( 1 \times 10^{-3} \) and a weight decay parameter of 0.05. Augmenting the training are methodologies such as label smoothing at 0.1 and the implementation of an exponential moving average (EMA). Subsequent to these specified techniques, no further training strategies are deployed.
 The VMamba-S did transfer training with pretrained weight get from \cite{43} on CNFOOD-241 with  a batch size of 32, else training strategies is the same as default VMamba.
 The CMAL-Net are trained from ResNet-50 pretrained weight from pytorch based on its original setting on github, the others models are trained with pretrained weights on ImageNet-1K loaded from Huggingface by timm module with an initial learning rate of \( 1 \times 10^{-4} \) and AdamW optimizer.

\subsection{Comparisons with State-of-the-Art Methods}
The experiment result shows that VMamba-based model are all surpass the latest SOTA methods with model pretrained on ImageNet-1k,

\subsection{Ablation Study on Res-VMamba }
To be fair to comparison Res-VMamba and VMamba, we both trained VMamba with and without the ImageNet-1k pretrained weight. For the computation resource issue, we do not pretrain Res-VMamba on the ImageNet-1k. In the result we observe that VMambs-S with ImageNet-1k pretrained weight can reach the SOTA on CNFOOD-241, and the Res-VMamba surpass the VMamba-S without pretrained weight on CNFOOD-241.

\hbadness=10000 
\begin{table*}
    \centering
    \begin{tabularx}{\linewidth}{@{}lXXXXXX@{}}
       \hline
        Model & Year & Use PW.? & Top-1 Val. Acc. & Top-5 Val. Acc. &Top-1 Test Acc. & Top-5 Test Acc.\\
        \hline
        VGG16  \cite{52}        & 2014 & Y & 66.98 & 90.10 & 65.06 & 89.60 \\
        ViT-B   \cite{15}       & 2020 & Y & 73.14 & 92.06 & 71.58 & 91.62 \\
        ResNet101 \cite{51}     & 2015 & Y & 74.42 & 93.62 & 72.59 & 93.16 \\
        DenseNet121  \cite{50}   & 2016 & Y & 76.46 & 94.57 & 74.77 & 94.29 \\
        Inceptionv4  \cite{49}  & 2016 & Y & 77.30 & 94.28 & 75.70 & 93.89 \\
        SEnet154  \cite{48}      & 2017 & Y & 77.47 & 94.86 & 76.02 & 94.61 \\
        PRENet   \cite{40}      & 2017 & Y*& 77.28 & 95.16 & 76.28 & 94.85 \\ 
        RepViT    \cite{47}     & 2023 & Y & 78.08 & 95.41 & 76.86 & 95.02 \\
        ConvNeXT-B \cite{46}    & 2022 & Y & 78.30 & 94.36 & 76.76 & 93.90 \\
        EfficientNet-B6 \cite{45}& 2019 & Y & 80.10 & 94.64 & 78.48 & 94.22 \\
        CMAL-Net \cite{44}       & 2023 & Y{$^\dagger$} & 80.16 & 95.99 & 78.56 & 95.40 \\
        \hline
        \textbf{VMamba-S \cite{43}}&\textbf{2024}&\textbf{N}&\textbf{79.17}&\textbf{95.64}&\textbf{77.73}&\textbf{95.24} \\
        \textbf{Res-VMamba (ours)}&\textbf{2024}&\textbf{N}&\textbf{79.54}&\textbf{95.72}&\textbf{78.26}&\textbf{95.31}\\ 
        \textbf{VMamba-S}&\textbf{2024}&\textbf{Y}{$^\ddagger$}&\textbf{82.15}&\textbf{96.91}&\textbf{80.58}&\textbf{96.71}\\
        \hline 
    \end{tabularx}
    \caption{Comparison of our approach (Res-Vmamba) to other baselines on CNFOOD-241 with our split method. PW., Val., Acc. denotes pretrained weight, validation, accuracy, respectively. *PRENet's pretrained weight got from official github is trained on Food2K dataset.  {$^\dagger$} CMAL-Net used the ResNet-50 pretrained weight from pytorch,  and other models' pretrained weights got from timm module loaded from Huggingface trained on ImageNet-1K dataset. {$^\ddagger$}VMamba's pretrained weight got from official github is trained on ImageNet-1K dataset.}
    \label{tab:my_label2}
\end{table*}
\hbadness=0 

\section{Discussion and Conclusion}
Fine-grained food classification poses a significant challenge in computer vision, and different food databases highlight various issues. In this study, we compared CNFOOD-241 and other food databases, revealing its characteristics as a high-resolution, data-imbalanced, and thus a challenging database. For instance, while PRENet achieves top-1 accuracies of 83.75\% on Food2K and 91.13\% on ETH Food-101, its top-1 accuracy on CNFOOD-241 is only 76.2\%, demonstrating the considerable difficulty of CNFOOD-241.
To address this challenge, we introduced the vision State Space Model, VMamba, in our research, which has superior performance compared to state-of-the-art (SOTA) models in fine-grained food recognition, with a notable improvement of 2.02\% in top-1 accuracy. Furthermore, our results indicated that incorporating a global residual architecture on VMamba (ResVMamba) can further enhance accuracy by 0.53\%.

Nevertheless, the current ResVMamba model has certain limitations. Due to resource constraints, we were unable to pre-train ResVMamba on ImageNet-1K. Additionally, the effectiveness of our global residual architecture when scaling up the VMamba model remains uncertain. Therefore, future research should continue exploring the classification capabilities of the VMamba model on a larger scale.

\section*{Acknowledgment}

Thanks Dr. Shao-Hsuan Chang and Kurh-Life Technology Co., Ltd. for providing a A100 cloud computing resource on Alicloud for the model training and fine-tuning.

\bibliographystyle{unsrtnat}
\bibliography{references}  

\begin{thebibliography}{71}
\providecommand{\natexlab}[1]{#1}
\providecommand{\url}[1]{\texttt{#1}}
\expandafter\ifx\csname urlstyle\endcsname\relax
  \providecommand{\doi}[1]{doi: #1}\else
  \providecommand{\doi}{doi: \begingroup \urlstyle{rm}\Url}\fi

\bibitem[Min et~al.(2019{\natexlab{a}})Min, Jiang, Liu, Rui, and Jain]{14}
Weiqing Min, Shuqiang Jiang, Linhu Liu, Yong Rui, and Ramesh Jain.
\newblock A survey on food computing.
\newblock \emph{ACM Comput. Surv.}, 52\penalty0 (5), sep 2019{\natexlab{a}}.
\newblock ISSN 0360-0300.
\newblock \doi{10.1145/3329168}.
\newblock URL \url{https://doi.org/10.1145/3329168}.

\bibitem[Wah et~al.(2011)Wah, Branson, Welinder, Perona, and Belongie]{1}
C.~Wah, S.~Branson, P.~Welinder, P.~Perona, and S.~Belongie.
\newblock Caltech-ucsd birds-200-2011 dataset.
\newblock Technical report, 2011.

\bibitem[Van~Horn et~al.(2015)Van~Horn, Branson, Farrell, Haber, Barry, Ipeirotis, Perona, and Belongie]{13}
Grant Van~Horn, Steve Branson, Ryan Farrell, Scott Haber, Jessie Barry, Panos Ipeirotis, Pietro Perona, and Serge Belongie.
\newblock Building a bird recognition app and large scale dataset with citizen scientists: The fine print in fine-grained dataset collection.
\newblock In \emph{2015 IEEE Conference on Computer Vision and Pattern Recognition (CVPR)}, pages 595--604, 2015.
\newblock \doi{10.1109/CVPR.2015.7298658}.

\bibitem[Zhou et~al.(2021)Zhou, Wang, Huang, Cui, and Shao]{2}
Yi~Zhou, Boyang Wang, Lei Huang, Shanshan Cui, and Ling Shao.
\newblock A benchmark for studying diabetic retinopathy: Segmentation, grading, and transferability.
\newblock \emph{IEEE Transactions on Medical Imaging}, 40\penalty0 (3):\penalty0 818--828, 2021.
\newblock \doi{10.1109/TMI.2020.3037771}.

\bibitem[Krause et~al.(2013)Krause, Stark, Deng, and Fei-Fei]{3}
Jonathan Krause, Michael Stark, Jia Deng, and Li~Fei-Fei.
\newblock 3d object representations for fine-grained categorization.
\newblock In \emph{2013 IEEE International Conference on Computer Vision Workshops}, pages 554--561, 2013.
\newblock \doi{10.1109/ICCVW.2013.77}.

\bibitem[Dong et~al.(2015)Dong, Wu, Pei, and Jia]{4}
Zhen Dong, Yuwei Wu, Mingtao Pei, and Yunde Jia.
\newblock Vehicle type classification using a semisupervised convolutional neural network.
\newblock \emph{IEEE Transactions on Intelligent Transportation Systems}, 16\penalty0 (4):\penalty0 2247--2256, 2015.
\newblock \doi{10.1109/TITS.2015.2402438}.

\bibitem[Maji et~al.(2013)Maji, Kannala, Rahtu, Blaschko, and Vedaldi]{5}
S.~Maji, J.~Kannala, E.~Rahtu, M.~Blaschko, and A.~Vedaldi.
\newblock Fine-grained visual classification of aircraft.
\newblock Technical report, 2013.

\bibitem[Dataset(2011)]{6}
E~Dataset.
\newblock Novel datasets for fine-grained image categorization.
\newblock In \emph{First Workshop on Fine Grained Visual Categorization, CVPR. Citeseer. Citeseer}. Citeseer, 2011.

\bibitem[Parkhi et~al.(2012)Parkhi, Vedaldi, Zisserman, and Jawahar]{7}
Omkar~M Parkhi, Andrea Vedaldi, Andrew Zisserman, and C.~V. Jawahar.
\newblock Cats and dogs.
\newblock In \emph{2012 IEEE Conference on Computer Vision and Pattern Recognition}, pages 3498--3505, 2012.
\newblock \doi{10.1109/CVPR.2012.6248092}.

\bibitem[Nilsback and Zisserman(2008)]{8}
Maria-Elena Nilsback and Andrew Zisserman.
\newblock Automated flower classification over a large number of classes.
\newblock In \emph{2008 Sixth Indian Conference on Computer Vision, Graphics \& Image Processing}, pages 722--729, 2008.
\newblock \doi{10.1109/ICVGIP.2008.47}.

\bibitem[Horn et~al.(2018)Horn, Aodha, Song, Cui, Sun, Shepard, Adam, Perona, and Belongie]{9}
G.~Van Horn, O.~Mac Aodha, Y.~Song, Y.~Cui, C.~Sun, A.~Shepard, H.~Adam, P.~Perona, and S.~Belongie.
\newblock The inaturalist species classification and detection dataset.
\newblock In \emph{2018 IEEE/CVF Conference on Computer Vision and Pattern Recognition (CVPR)}, pages 8769--8778, Los Alamitos, CA, USA, jun 2018. IEEE Computer Society.
\newblock \doi{10.1109/CVPR.2018.00914}.
\newblock URL \url{https://doi.ieeecomputersociety.org/10.1109/CVPR.2018.00914}.

\bibitem[Horn et~al.(2021)Horn, Cole, Beery, Wilber, Belongie, and MacAodha]{10}
G.~Van Horn, E.~Cole, S.~Beery, K.~Wilber, S.~Belongie, and O.~MacAodha.
\newblock Benchmarking representation learning for natural world image collections.
\newblock In \emph{2021 IEEE/CVF Conference on Computer Vision and Pattern Recognition (CVPR)}, pages 12879--12888, Los Alamitos, CA, USA, jun 2021. IEEE Computer Society.
\newblock \doi{10.1109/CVPR46437.2021.01269}.
\newblock URL \url{https://doi.ieeecomputersociety.org/10.1109/CVPR46437.2021.01269}.

\bibitem[Singla et~al.(2016)Singla, Yuan, and Ebrahimi]{11}
Ashutosh Singla, Lin Yuan, and Touradj Ebrahimi.
\newblock Food/non-food image classification and food categorization using pre-trained googlenet model.
\newblock In \emph{Proceedings of the 2nd International Workshop on Multimedia Assisted Dietary Management}, MADiMa '16, page 3–11, New York, NY, USA, 2016. Association for Computing Machinery.
\newblock ISBN 9781450345200.
\newblock \doi{10.1145/2986035.2986039}.
\newblock URL \url{https://doi.org/10.1145/2986035.2986039}.

\bibitem[Fan(2022)]{12}
Bokun Fan.
\newblock Cnfood-241.
\newblock Mendeley Data, 2022.
\newblock \doi{10.17632/fspyss5zbb.1}.

\bibitem[Liu et~al.(2024{\natexlab{a}})Liu, Tian, Zhao, Yu, Xie, Wang, Ye, and Liu]{43}
Y.~Liu, Y.~Tian, Y.~Zhao, H.~Yu, L.~Xie, Y.~Wang, Q.~Ye, and Y.~Liu.
\newblock Vmamba: Visual state space model.
\newblock Technical report, 2024{\natexlab{a}}.

\bibitem[Deng et~al.(2009)Deng, Dong, Socher, Li, Li, and Fei-Fei]{42}
Jia Deng, Wei Dong, Richard Socher, Li-Jia Li, Kai Li, and Li~Fei-Fei.
\newblock Imagenet: A large-scale hierarchical image database.
\newblock In \emph{2009 IEEE conference on computer vision and pattern recognition}, pages 248--255. Ieee, 2009.

\bibitem[Fan et~al.(2023)Fan, Li, Dong, Li, and Nie]{65}
Bokun Fan, Weiqi Li, Liang Dong, Jingzhen Li, and Zedong Nie.
\newblock Automatic chinese food recognition based on a stacking fusion model.
\newblock In \emph{Annu Int Conf IEEE Eng Med Biol Soc.}, 2023.
\newblock \doi{10.1109/EMBC40787.2023.10340620}.

\bibitem[Chen et~al.(2017)Chen, Zhou, Zhu, and Diao]{32}
Xin Chen, Hua Zhou, Yu~Zhu, and Liang Diao.
\newblock Chinesefoodnet: A large-scale image dataset for chinese food recognition.
\newblock \emph{arXiv preprint arXiv:1705.02743}, 2017.

\bibitem[Chen et~al.(2009)Chen, Dhingra, Wu, Yang, Sukthankar, and Yang]{16}
Mei Chen, Kapil Dhingra, Wen Wu, Lei Yang, Rahul Sukthankar, and Jie Yang.
\newblock Pfid: Pittsburgh fast-food image dataset.
\newblock In \emph{2009 16th IEEE International Conference on Image Processing (ICIP)}, pages 289--292, 2009.
\newblock \doi{10.1109/ICIP.2009.5413511}.

\bibitem[Joutou and Yanai(2009{\natexlab{a}})]{17}
Taichi Joutou and Keiji Yanai.
\newblock A food image recognition system with multiple kernel learning.
\newblock In \emph{2009 16th IEEE International Conference on Image Processing (ICIP)}, pages 285--288, 2009{\natexlab{a}}.
\newblock \doi{10.1109/ICIP.2009.5413400}.

\bibitem[Hoashi et~al.(2010)Hoashi, Joutou, and Yanai]{18}
Hajime Hoashi, Taichi Joutou, and Keiji Yanai.
\newblock Image recognition of 85 food categories by feature fusion.
\newblock In \emph{2010 IEEE International Symposium on Multimedia}, pages 296--301, 2010.
\newblock \doi{10.1109/ISM.2010.51}.

\bibitem[Matsuda and Yanai(2012)]{19}
Yuji Matsuda and Keiji Yanai.
\newblock Multiple-food recognition considering co-occurrence employing manifold ranking.
\newblock In \emph{Proceedings of the 21st International Conference on Pattern Recognition (ICPR2012)}, pages 2017--2020, 2012.

\bibitem[Kawano and Yanai(2015)]{20}
Yoshiyuki Kawano and Keiji Yanai.
\newblock Automatic expansion of a food image dataset leveraging existing categories with domain adaptation.
\newblock In Lourdes Agapito, Michael~M. Bronstein, and Carsten Rother, editors, \emph{Computer Vision - ECCV 2014 Workshops}, pages 3--17, Cham, 2015. Springer International Publishing.
\newblock ISBN 978-3-319-16199-0.
\newblock \doi{10.1007/978-3-319-16199-0_1}.

\bibitem[Bossard et~al.(2014)Bossard, Guillaumin, and Van~Gool]{21}
Lukas Bossard, Matthieu Guillaumin, and Luc Van~Gool.
\newblock Food-101 -- mining discriminative components with random forests.
\newblock In David Fleet, Tomas Pajdla, Bernt Schiele, and Tinne Tuytelaars, editors, \emph{Computer Vision -- ECCV 2014}, pages 446--461, Cham, 2014. Springer International Publishing.
\newblock ISBN 978-3-319-10599-4.

\bibitem[Anthimopoulos et~al.(2014)Anthimopoulos, Gianola, Scarnato, Diem, and Mougiakakou]{22}
Marios~M. Anthimopoulos, Lauro Gianola, Luca Scarnato, Peter Diem, and Stavroula~G. Mougiakakou.
\newblock A food recognition system for diabetic patients based on an optimized bag-of-features model.
\newblock \emph{IEEE Journal of Biomedical and Health Informatics}, 18\penalty0 (4):\penalty0 1261--1271, 2014.
\newblock \doi{10.1109/JBHI.2014.2308928}.

\bibitem[Wang et~al.(2015)Wang, Kumar, Thome, Cord, and Precioso]{23}
Xin Wang, Devinder Kumar, Nicolas Thome, Matthieu Cord, and Frédéric Precioso.
\newblock Recipe recognition with large multimodal food dataset.
\newblock In \emph{2015 IEEE International Conference on Multimedia \& Expo Workshops (ICMEW)}, pages 1--6, 2015.
\newblock \doi{10.1109/ICMEW.2015.7169757}.

\bibitem[Xu et~al.(2015)Xu, Herranz, Jiang, Wang, Song, and Jain]{24}
Ruihan Xu, Luis Herranz, Shuqiang Jiang, Shuang Wang, Xinhang Song, and Ramesh Jain.
\newblock Geolocalized modeling for dish recognition.
\newblock \emph{IEEE Transactions on Multimedia}, 17\penalty0 (8):\penalty0 1187--1199, 2015.
\newblock \doi{10.1109/TMM.2015.2438717}.

\bibitem[Farinella et~al.(2015)Farinella, Allegra, and Stanco]{25}
Giovanni~Maria Farinella, Dario Allegra, and Filippo Stanco.
\newblock A benchmark dataset to study the representation of food images.
\newblock In Lourdes Agapito, Michael~M. Bronstein, and Carsten Rother, editors, \emph{Computer Vision - ECCV 2014 Workshops}, pages 584--599, Cham, 2015. Springer International Publishing.
\newblock ISBN 978-3-319-16199-0.

\bibitem[Chen and Ngo(2016)]{26}
Jingjing Chen and Chong-wah Ngo.
\newblock Deep-based ingredient recognition for cooking recipe retrieval.
\newblock In \emph{Proceedings of the 24th ACM International Conference on Multimedia}, MM '16, page 32–41, New York, NY, USA, 2016. Association for Computing Machinery.
\newblock ISBN 9781450336031.
\newblock \doi{10.1145/2964284.2964315}.
\newblock URL \url{https://doi.org/10.1145/2964284.2964315}.

\bibitem[Zhou and Lin(2016)]{27}
Feng Zhou and Yuanqing Lin.
\newblock Fine-grained image classification by exploring bipartite-graph labels.
\newblock In \emph{2016 IEEE Conference on Computer Vision and Pattern Recognition (CVPR)}, pages 1124--1133, 2016.
\newblock \doi{10.1109/CVPR.2016.127}.

\bibitem[Merler et~al.(2016)Merler, Wu, Uceda-Sosa, Nguyen, and Smith]{28}
Michele Merler, Hui Wu, Rosario Uceda-Sosa, Quoc-Bao Nguyen, and John~R. Smith.
\newblock Snap, eat, repeat: A food recognition engine for dietary logging.
\newblock In \emph{Proceedings of the 2nd International Workshop on Multimedia Assisted Dietary Management}, MADiMa '16, page 31–40, New York, NY, USA, 2016. Association for Computing Machinery.
\newblock ISBN 9781450345200.
\newblock \doi{10.1145/2986035.2986036}.
\newblock URL \url{https://doi.org/10.1145/2986035.2986036}.

\bibitem[Farinella et~al.(2016)Farinella, Allegra, Moltisanti, Stanco, and Battiato]{30}
Giovanni~Maria Farinella, Dario Allegra, Marco Moltisanti, Filippo Stanco, and Sebastiano Battiato.
\newblock Retrieval and classification of food images.
\newblock \emph{Computers in Biology and Medicine}, 77:\penalty0 23--39, 2016.
\newblock ISSN 0010-4825.
\newblock \doi{https://doi.org/10.1016/j.compbiomed.2016.07.006}.
\newblock URL \url{https://www.sciencedirect.com/science/article/pii/S0010482516301822}.

\bibitem[Ciocca et~al.(2017)Ciocca, Napoletano, and Schettini]{31}
Gianluigi Ciocca, Paolo Napoletano, and Raimondo Schettini.
\newblock Learning cnn-based features for retrieval of food images.
\newblock In Sebastiano Battiato, Giovanni~Maria Farinella, Marco Leo, and Giovanni Gallo, editors, \emph{New Trends in Image Analysis and Processing -- ICIAP 2017}, pages 426--434, Cham, 2017. Springer International Publishing.
\newblock ISBN 978-3-319-70742-6.

\bibitem[Hou et~al.(2017)Hou, Feng, and Wang]{33}
Saihui Hou, Yushan Feng, and Zilei Wang.
\newblock Vegfru: A domain-specific dataset for fine-grained visual categorization.
\newblock In \emph{2017 IEEE International Conference on Computer Vision (ICCV)}, pages 541--549, 2017.
\newblock \doi{10.1109/ICCV.2017.66}.

\bibitem[Qiu et~al.(2019)Qiu, P.-W.~Lo, Sun, Wang, and Lo]{34}
Jianing Qiu, Frank P.-W.~Lo, Yingnan Sun, Siyao Wang, and Benny Lo.
\newblock Mining discriminative food regions for accurate food recognition.
\newblock In \emph{BMVC}, 2019.

\bibitem[Kaur et~al.(2019)Kaur, Sikka, Wang, Belongie, and Divakaran]{35}
P.~Kaur, K.~Sikka, W.~Wang, S.~Belongie, and A.~Divakaran.
\newblock Foodx-251: a dataset for fine-grained food classification.
\newblock 2019.
\newblock \doi{10.48550/arxiv.1907.06167}.

\bibitem[Min et~al.(2019{\natexlab{b}})Min, Liu, Luo, and Jiang]{36}
Weiqing Min, Linhu Liu, Zhengdong Luo, and Shuqiang Jiang.
\newblock Ingredient-guided cascaded multi-attention network for food recognition.
\newblock In \emph{Proceedings of the 27th ACM International Conference on Multimedia}, MM '19, page 1331–1339, New York, NY, USA, 2019{\natexlab{b}}. Association for Computing Machinery.
\newblock ISBN 9781450368896.
\newblock \doi{10.1145/3343031.3350948}.
\newblock URL \url{https://doi.org/10.1145/3343031.3350948}.

\bibitem[Sahoo et~al.(2019)Sahoo, Hao, Ke, Xiongwei, Le, Achananuparp, Lim, and Hoi]{37}
Doyen Sahoo, Wang Hao, Shu Ke, Wu~Xiongwei, Hung Le, Palakorn Achananuparp, Ee-Peng Lim, and Steven C.~H. Hoi.
\newblock Foodai: Food image recognition via deep learning for smart food logging.
\newblock In \emph{Proceedings of the 25th ACM SIGKDD International Conference on Knowledge Discovery \& Data Mining}, KDD '19, page 2260–2268, New York, NY, USA, 2019. Association for Computing Machinery.
\newblock ISBN 9781450362016.
\newblock \doi{10.1145/3292500.3330734}.
\newblock URL \url{https://doi.org/10.1145/3292500.3330734}.

\bibitem[Yang(2020)]{38}
Tsan-Lun Yang.
\newblock Taiwanese-food-101.
\newblock Technical report, 2020.
\newblock URL \url{https://ieeexplore.ieee.org/document/708428}.
\newblock Retrieved Jan. 3, 2024.

\bibitem[Min et~al.(2020)Min, Liu, Wang, Luo, Wei, Wei, and Jiang]{39}
Weiqing Min, Linhu Liu, Zhiling Wang, Zhengdong Luo, Xiaoming Wei, Xiaolin Wei, and Shuqiang Jiang.
\newblock Isia food-500: A dataset for large-scale food recognition via stacked global-local attention network.
\newblock In \emph{Proceedings of the 28th ACM International Conference on Multimedia}, 2020.

\bibitem[Min et~al.(2023)Min, Wang, Liu, Luo, Kang, Wei, Wei, and Jiang]{40}
Weiqing Min, Zhiling Wang, Yuxin Liu, Mengjiang Luo, Liping Kang, Xiaoming Wei, Xiaolin Wei, and Shuqiang Jiang.
\newblock Large scale visual food recognition.
\newblock \emph{IEEE Transactions on Pattern Analysis and Machine Intelligence}, 45\penalty0 (8):\penalty0 9932--9949, 2023.
\newblock \doi{10.1109/TPAMI.2023.3237871}.

\bibitem[SP et~al.(2022)SP, G, EA, D, H, V, and M.]{41}
Mohanty SP, Singhal G, Scuccimarra EA, Kebaili D, Héritier H, Boulanger V, and Salathé M.
\newblock The food recognition benchmark: Using deep learning to recognize food in images.
\newblock \emph{Front Nutr.}, 2022.
\newblock \doi{10.3389/fnut.2022.875143}.

\bibitem[Liu et~al.(2023)Liu, Zhao, Wang, and Kato]{44}
Dichao Liu, Longjiao Zhao, Yu~Wang, and Jien Kato.
\newblock Learn from each other to classify better: Cross-layer mutual attention learning for fine-grained visual classification.
\newblock \emph{Pattern Recognition}, 140:\penalty0 109550, 2023.
\newblock ISSN 0031-3203.
\newblock \doi{https://doi.org/10.1016/j.patcog.2023.109550}.
\newblock URL \url{https://www.sciencedirect.com/science/article/pii/S0031320323002509}.

\bibitem[Mingxing~Tan(2019)]{45}
Quoc V.~Le Mingxing~Tan.
\newblock Efficientnet: Rethinking model scaling for convolutional neural networks.
\newblock Technical report, 2019.

\bibitem[Liu et~al.(2022)Liu, Mao, Wu, Feichtenhofer, Darrell, and Xie]{46}
Zhuang Liu, Hanzi Mao, Chao-Yuan Wu, Christoph Feichtenhofer, Trevor Darrell, and Saining Xie.
\newblock A convnet for the 2020s.
\newblock Technical report, 2022.

\bibitem[Liu et~al.(2021)Liu, Lin, Cao, Hu, Wei, Zhang, Lin, and Guo]{53}
Z.~Liu, Y.~Lin, Y.~Cao, H.~Hu, Y.~Wei, Z.~Zhang, S.~Lin, and B.~Guo.
\newblock Swin transformer: Hierarchical vision transformer using shifted windows.
\newblock In \emph{2021 IEEE/CVF International Conference on Computer Vision (ICCV)}, pages 9992--10002, Los Alamitos, CA, USA, oct 2021. IEEE Computer Society.
\newblock \doi{10.1109/ICCV48922.2021.00986}.
\newblock URL \url{https://doi.ieeecomputersociety.org/10.1109/ICCV48922.2021.00986}.

\bibitem[Wang et~al.(2023)Wang, Chen, Lin, Han, and Ding]{47}
Ao~Wang, Hui Chen, Zijia Lin, Jungong Han, and Guiguang Ding.
\newblock Repvit: Revisiting mobile cnn from vit perspective.
\newblock Technical report, 2023.

\bibitem[Hu et~al.(2018)Hu, Shen, and Sun]{48}
Jie Hu, Li~Shen, and Gang Sun.
\newblock Squeeze-and-excitation networks.
\newblock In \emph{2018 IEEE/CVF Conference on Computer Vision and Pattern Recognition}, pages 7132--7141, 2018.
\newblock \doi{10.1109/CVPR.2018.00745}.

\bibitem[Joutou and Yanai(2009{\natexlab{b}})]{67}
Taichi Joutou and Keiji Yanai.
\newblock A food image recognition system with multiple kernel learning.
\newblock In \emph{2009 16th IEEE International Conference on Image Processing (ICIP)}, pages 285--288, 2009{\natexlab{b}}.
\newblock \doi{10.1109/ICIP.2009.5413400}.

\bibitem[Ravì et~al.(2015)Ravì, Lo, and Yang]{66}
Daniele Ravì, Benny Lo, and Guang-Zhong Yang.
\newblock Real-time food intake classification and energy expenditure estimation on a mobile device.
\newblock In \emph{2015 IEEE 12th International Conference on Wearable and Implantable Body Sensor Networks (BSN)}, pages 1--6, 2015.
\newblock \doi{10.1109/BSN.2015.7299410}.

\bibitem[Hearst et~al.(1998)Hearst, Dumais, Osuna, Platt, and Scholkopf]{72}
M.A. Hearst, S.T. Dumais, E.~Osuna, J.~Platt, and B.~Scholkopf.
\newblock Support vector machines.
\newblock \emph{IEEE Intelligent Systems and their Applications}, 13\penalty0 (4):\penalty0 18--28, 1998.
\newblock \doi{10.1109/5254.708428}.

\bibitem[Rahmat and Kutty(2021)]{68}
Rafhan~Amnani Rahmat and Suhaili~Beeran Kutty.
\newblock Malaysian food recognition using alexnet cnn and transfer learning.
\newblock In \emph{2021 IEEE 11th IEEE Symposium on Computer Applications \& Industrial Electronics (ISCAIE)}, pages 59--64, 2021.
\newblock \doi{10.1109/ISCAIE51753.2021.9431833}.

\bibitem[Zahisham et~al.(2020)Zahisham, Lee, and Lim]{69}
Zharfan Zahisham, Chin~Poo Lee, and Kian~Ming Lim.
\newblock Food recognition with resnet-50.
\newblock In \emph{2020 IEEE 2nd International Conference on Artificial Intelligence in Engineering and Technology (IICAIET)}, pages 1--5, 2020.
\newblock \doi{10.1109/IICAIET49801.2020.9257825}.

\bibitem[G. et~al.(2022)G., Vutkur, and P.]{70}
VijayaKumari G., Priyanka Vutkur, and Vishwanath P.
\newblock Food classification using transfer learning technique.
\newblock \emph{Global Transitions Proceedings}, 3\penalty0 (1):\penalty0 225--229, 2022.
\newblock ISSN 2666-285X.
\newblock \doi{https://doi.org/10.1016/j.gltp.2022.03.027}.
\newblock URL \url{https://www.sciencedirect.com/science/article/pii/S2666285X22000334}.
\newblock International Conference on Intelligent Engineering Approach(ICIEA-2022).

\bibitem[Hassannejad et~al.(2016)Hassannejad, Matrella, Ciampolini, De~Munari, Mordonini, and Cagnoni]{71}
Hamid Hassannejad, Guido Matrella, Paolo Ciampolini, Ilaria De~Munari, Monica Mordonini, and Stefano Cagnoni.
\newblock Food image recognition using very deep convolutional networks.
\newblock In \emph{Proceedings of the 2nd International Workshop on Multimedia Assisted Dietary Management}, MADiMa '16, page 41–49, New York, NY, USA, 2016. Association for Computing Machinery.
\newblock ISBN 9781450345200.
\newblock \doi{10.1145/2986035.2986042}.
\newblock URL \url{https://doi.org/10.1145/2986035.2986042}.

\bibitem[Gu et~al.(2022)Gu, Goel, and Re]{54}
Albert Gu, Karan Goel, and Christopher Re.
\newblock Efficiently modeling long sequences with structured state spaces.
\newblock In \emph{International Conference on Learning Representations}, 2022.
\newblock URL \url{https://openreview.net/forum?id=uYLFoz1vlAC}.

\bibitem[Albert~Gu(2023)]{55}
Tri~Dao Albert~Gu.
\newblock Mamba: Linear-time sequence modeling with selective state spaces.
\newblock Technical report, 2023.

\bibitem[Zhu et~al.(2024)Zhu, Liao, Zhang, Wang, Liu, and Wang]{56}
Lianghui Zhu, Bencheng Liao, Qian Zhang, Xinlong Wang, Wenyu Liu, and Xinggang Wang.
\newblock Vision mamba: Efficient visual representation learning with bidirectional state space model.
\newblock Technical report, 2024.

\bibitem[Jun~Ma(2024)]{57}
Bo~Wang Jun~Ma, Feifei~Li.
\newblock U-mamba: Enhancing long-range dependency for biomedical image segmentation.
\newblock Technical report, 2024.

\bibitem[Jiacheng~Ruan(2024)]{58}
Suncheng~Xiang Jiacheng~Ruan.
\newblock Vm-unet: Vision mamba unet for medical image segmentation.
\newblock Technical report, 2024.

\bibitem[Liu et~al.(2024{\natexlab{b}})Liu, Yang, Zhou, Xi, Yu, Yu, Liang, Shi, Zhang, Zheng, and Wang]{59}
Jiarun Liu, Hao Yang, Hong-Yu Zhou, Yan Xi, Lequan Yu, Yizhou Yu, Yong Liang, Guangming Shi, Shaoting Zhang, Hairong Zheng, and Shanshan Wang.
\newblock Swin-umamba: Mamba-based unet with imagenet-based pretraining.
\newblock Technical report, 2024{\natexlab{b}}.

\bibitem[Gong et~al.(2024)Gong, Kang, Wang, Wan, and Li]{60}
Haifan Gong, Luoyao Kang, Yitao Wang, Xiang Wan, and Haofeng Li.
\newblock nnmamba: 3d biomedical image segmentation, classification and landmark detection with state space model.
\newblock Technical report, 2024.

\bibitem[Tao~Guo(2024)]{61}
Cai~Meng Tao~Guo, Yinuo~Wang.
\newblock Mambamorph: a mamba-based backbone with contrastive feature learning for deformable mr-ct registration.
\newblock Technical report, 2024.

\bibitem[Yijun~Yang(2024)]{62}
Lei~Zhu Yijun~Yang, Zhaohu~Xing.
\newblock Vivim: a video vision mamba for medical video object segmentation.
\newblock Technical report, 2024.

\bibitem[He et~al.(2016)He, Zhang, Ren, and Sun]{51}
Kaiming He, Xiangyu Zhang, Shaoqing Ren, and Jian Sun.
\newblock Deep residual learning for image recognition.
\newblock In \emph{2016 IEEE Conference on Computer Vision and Pattern Recognition (CVPR)}, pages 770--778, 2016.
\newblock \doi{10.1109/CVPR.2016.90}.

\bibitem[Gu et~al.(2021)Gu, Johnson, Goel, Saab, Dao, Rudra, and Re]{63}
Albert Gu, Isys Johnson, Karan Goel, Khaled~Kamal Saab, Tri Dao, Atri Rudra, and Christopher Re.
\newblock Combining recurrent, convolutional, and continuous-time models with linear state space layers.
\newblock In A.~Beygelzimer, Y.~Dauphin, P.~Liang, and J.~Wortman Vaughan, editors, \emph{Advances in Neural Information Processing Systems}, 2021.
\newblock URL \url{https://openreview.net/forum?id=yWd42CWN3c}.

\bibitem[Gupta et~al.(2022)Gupta, Gu, and Berant]{64}
Ankit Gupta, Albert Gu, and Jonathan Berant.
\newblock Diagonal state spaces are as effective as structured state spaces.
\newblock In Alice~H. Oh, Alekh Agarwal, Danielle Belgrave, and Kyunghyun Cho, editors, \emph{Advances in Neural Information Processing Systems}, 2022.
\newblock URL \url{https://openreview.net/forum?id=RjS0j6tsSrf}.

\bibitem[Simonyan and Zisserman(2015)]{52}
K~Simonyan and A~Zisserman.
\newblock Very deep convolutional networks for large-scale image recognition.
\newblock pages 1--14. Computational and Biological Learning Society, 2015.

\bibitem[Dosovitskiy et~al.(2021)Dosovitskiy, Beyer, Kolesnikov, Weissenborn, Zhai, Unterthiner, Dehghani, Minderer, Heigold, Gelly, Uszkoreit, and Houlsby]{15}
Alexey Dosovitskiy, Lucas Beyer, Alexander Kolesnikov, Dirk Weissenborn, Xiaohua Zhai, Thomas Unterthiner, Mostafa Dehghani, Matthias Minderer, Georg Heigold, Sylvain Gelly, Jakob Uszkoreit, and Neil Houlsby.
\newblock An image is worth 16x16 words: Transformers for image recognition at scale.
\newblock In \emph{International Conference on Learning Representations}, 2021.
\newblock URL \url{https://openreview.net/forum?id=YicbFdNTTy}.

\bibitem[Huang et~al.(2017)Huang, Liu, Maaten, and Weinberger]{50}
G.~Huang, Z.~Liu, L.~Van~Der Maaten, and K.~Q. Weinberger.
\newblock Densely connected convolutional networks.
\newblock In \emph{2017 IEEE Conference on Computer Vision and Pattern Recognition (CVPR)}, pages 2261--2269, Los Alamitos, CA, USA, jul 2017. IEEE Computer Society.
\newblock \doi{10.1109/CVPR.2017.243}.
\newblock URL \url{https://doi.ieeecomputersociety.org/10.1109/CVPR.2017.243}.

\bibitem[Szegedy et~al.(2016)Szegedy, Ioffe, Vanhoucke, and Alemi]{49}
Christian Szegedy, Sergey Ioffe, Vincent Vanhoucke, and Alex Alemi.
\newblock Inception-v4, inception-resnet and the impact of residual connections on learning.
\newblock Technical report, 2016.

\end{thebibliography}






\end{document}